\documentclass{article}

\usepackage[numbers]{natbib}

\usepackage[utf8]{inputenc} 
\usepackage[T1]{fontenc}    
\usepackage{hyperref}       
\usepackage{url}            
\usepackage{booktabs}       
\usepackage{amsfonts}       
\usepackage{nicefrac}       
\usepackage{microtype}      
\usepackage{xcolor}         

\usepackage{graphicx}
\usepackage[ruled,vlined]{algorithm2e}
\usepackage{subfig}
\usepackage{bbm}
\usepackage{gensymb}

\usepackage{multirow}
\usepackage{booktabs}
\usepackage{amsmath}

\usepackage{amsmath,amsfonts,bm}

\def\eqref#1{equation~\ref{#1}}

\def\1{\bm{1}}

\DeclareMathAlphabet{\mathsfit}{\encodingdefault}{\sfdefault}{m}{sl}
\SetMathAlphabet{\mathsfit}{bold}{\encodingdefault}{\sfdefault}{bx}{n}

\title{AI Empowered Net-RCA for 6G}

\author{Chengbo Qiu, Kai Yang, Ji Wang, and Shenjie Zhao
}

\begin{document}

\maketitle

\begin{abstract}
6G is envisioned to offer higher data rate, improved reliability, ubiquitous AI services, and support massive scale of connected devices. As a consequence, 6G will be much more complex than its predecessors. The growth of the system scale and complexity as well as the coexistence with the legacy networks and the diversified service requirements will inevitably incur huge maintenance cost and efforts for future 6G networks. Network Root Cause Analysis (Net-RCA) plays a critical role in identifying root causes of network faults. In this article, we first give an introduction about the envisioned 6G networks. Next, we discuss the challenges and potential solutions of 6G network operation and management, and comprehensively survey existing RCA methods. Then we propose an artificial intelligence (AI)-empowered Net-RCA framework for 6G. Performance comparisons on both synthetic and real-world network data are carried out to demonstrate that the proposed method outperforms the existing method considerably.
\end{abstract}
 
\section{Introduction}

In order to realize the vision of connecting everything worldwide, the sixth-generation (6G) wireless networks are receiving unprecedented attention, and are anticipated to build a bridge to the smart society of the future. Compared with 5G, 6G is expected to boost network spectrum efficiency, offer massive access, improved reliability, and latency, as shown in Table \ref{tab:table1}. Consequently, future 6G networks are expected to be able to support wireless connections of various emerging applications and massive intelligent devices e.g., extended reality (XR) services, telemedicine and brain-computer interfaces, and deliver low latency and high data rates for different heterogeneous devices. We illustrate the key emerging 6G applications in Fig.1. Specifically, the application types of 6G can be classified as MBRLLC, mURLLC, HCS and MPS \cite{bib1}. The most representative 6G use cases are presented as follow, the network requirements are shown in Table \ref{tab:table2}.

\begin{figure*}[!t]
\centering
\includegraphics[width=1.0\textwidth]{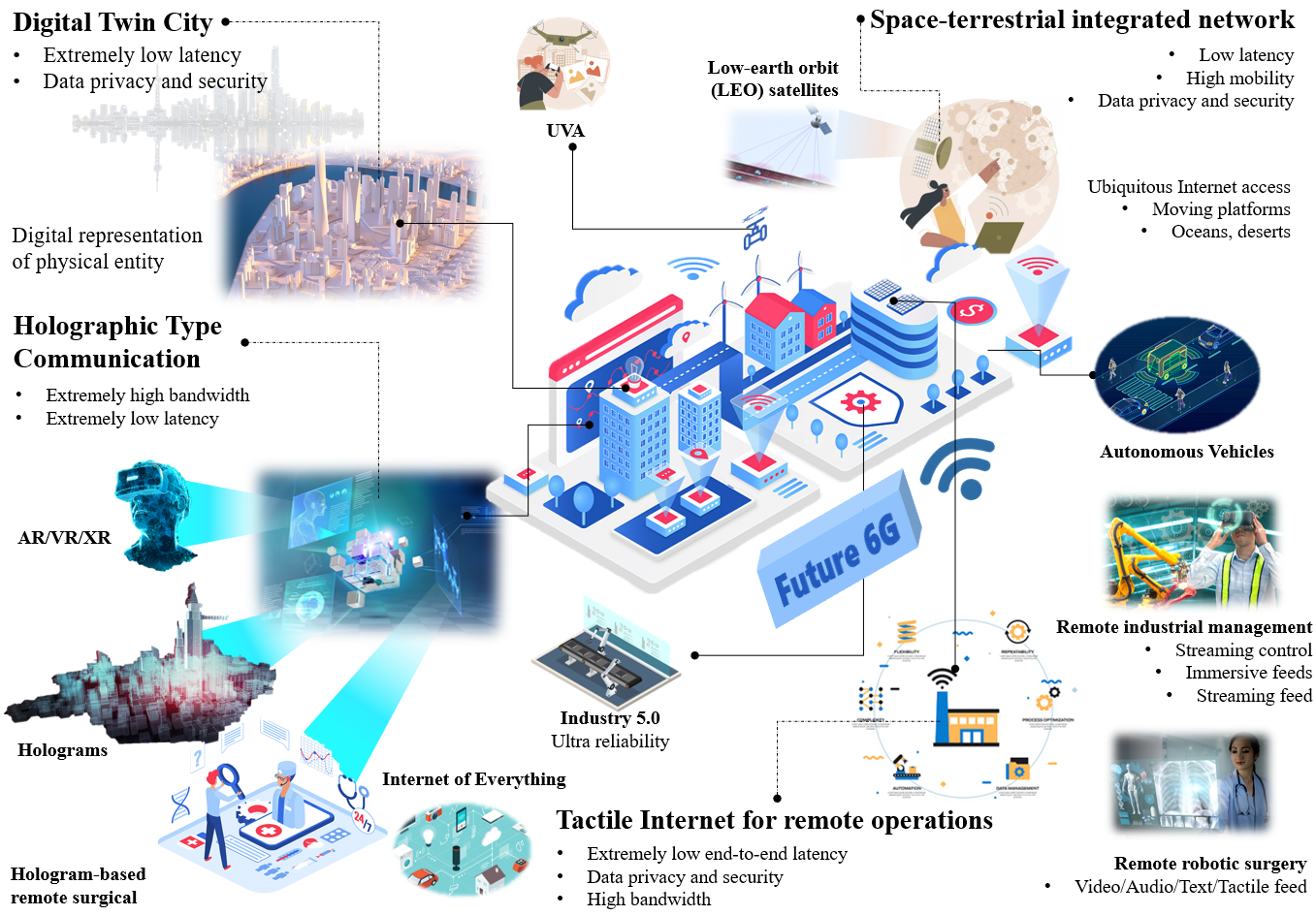}
\caption{Emerging 6G applications.}
\label{fig_1}
\end{figure*}

\textit{Holographic Type Communication (HTC)}: HTC is expected to deliver 3D images from one or multiple source nodes to different destinations. Owing to the extremely large data for recording and reconstructing, HTC requires bandwidth up to Tbps level for transmission.

\textit{Tactile Internet of Remote Operation (TIRO)}: TIRO can realize real-time control and immerse interaction of remote infrastructures. For example, remote robotic surgery requires surgeons to perform surgical actions based on the real-time visual and haptic feedbacks. Thus, TIRO requires ultralow network latency for real-time interaction, and high bandwidth for various feedbacks.

\textit{Intelligent Operation Network (ION)}: ION aims to employ intelligent technologies to make a comprehensive, multi-level, and deeply correlated analysis of historical and real-time measured data, and establish an intelligent framework to diagnose network faults and invoke automatic recovery mechanism. ION requires real-time data collection with low latency.

\textit{Network and Computing Convergence (NCC)}: NCC aims to realize intelligent load-balancing among multiple edge sites through unified management, control and operation of network capabilities.

\textit{Digital Twin (DT)}: By mapping various scenarios with physical objects to digital world, DT can provide better responses for predictive maintenance and optimization. For example, a digital twin city is a smart city framework based on DT paradigm, which supports forecasting of changes in the state of urban infrastructure by interconnecting physical city with virtualized city.

\textit{Satellite-Terrestrial Integrated Network (STIN)}: STIN is a scenario of future seamlessly integrated space and terrestrial Internet framework, where devices can realize ubiquitous worldwide internet access, including remote areas such as oceans, deserts or moving planes.

\textit{Industrial IoT (IIoT) with cloudification}: As traditional control functions are slowly virtualized and moved from hardware to the cloud, IIoT with cloudification is designed to realize automatic operation and control of industrial processes and provides factory-wide, as well as inter-factory large-scale connectivity in the future. It requires higher level of latency, security and reliability.

\renewcommand\arraystretch{1.5}
\begin{table*}[!ht]
\small
\caption{Comparisons of Key Performances, Network Characteristics and Application Types between 4G, 5G, and 6G. \label{tab:table1}}
\centering
\resizebox{\linewidth}{!}{
\begin{tabular}{lccc}
\bottomrule
                                                           & \textbf{4G Networks}                                            & \textbf{5G Networks}                                                                                                                                                                                       & \textbf{6G Networks}                                                                                                                                                                                                                           \\ \hline
\textbf{Peak data rate}                                    & 1 Gbps                                                 & 20 Gbps                                                                                                                                                                                           & $\geq$1 Tbps                                                                                                                                                                                                                               \\ \cline{2-4} 
\textbf{User-experienced data rate}                        & 10 Mbps                                                & 100 Mbps                                                                                                                                                                                          & 1Gbps                                                                                                                                                                                                                                 \\ \cline{2-4} 
\textbf{Spectrum efficiency}                               & 1×                                                     & 3×                                                                                                                                                                                                & 15-30×                                                                                                                                                                                                                                \\ \cline{2-4} 
\textbf{Network energy efficiency}                         & 1×                                                     & 10×                                                                                                                                                                                               & 100×                                                                                                                                                                                                                                  \\ \cline{2-4} 
\textbf{Area traffic capacity}                             & 0.1Mb/s/m$^{2}$                           & 10Mb/s/m$^{2}$                                                                                                                                                                       & $\geq$1Gb/s/m$^{2}$                                                                                                                                                                                                           \\ \cline{2-4} 
\textbf{Connection density (device/km$^{2}$)} & 10$^{5}$                                  & 10$^{6}$                                                                                                                                                                             & 10$^{7}$                                                                                                                                                                                                                 \\ \cline{2-4} 
\textbf{Latency (ms)}                                      & 10                                                     & 1                                                                                                                                                                                                 & 0.01-0.1                                                                                                                                                                                                                              \\ \cline{2-4} 
\textbf{Mobility (km/h)}                                   & 350                                                    & 500                                                                                                                                                                                               & 1000                                                                                                                                                                                                                                  \\ \cline{2-4} 
\textbf{End-to-end reliability (percent)}                  & -                                                      & 99.999                                                                                                                                                                                            & 99.99999                                                                                                                                                                                                                              \\ \cline{2-4} 
\textbf{Network characteristics}                           & \begin{tabular}[c]{@{}c@{}}Flat,\\ All IP\end{tabular} & \begin{tabular}[c]{@{}c@{}}Cloudization,\\ Softwarization,\\ Virtualiszation,\\ Slicing\end{tabular}                                                                                               & \begin{tabular}[c]{@{}c@{}}Cloudization,\\ Softwarization,\\ Virtualiszation,\\ Slicing,\\ Intelligentization\end{tabular}                                                                                                             \\ \cline{2-4}
\textbf{Application types}                                 & \multicolumn{1}{l}{MBB (mobile broadband)}            & \multicolumn{1}{l}{\begin{tabular}[c]{@{}l@{}}eMBB (Enhanced Mobile\\Broadband), \\ URLLC (ultra-reliable low\\latency communications), \\ mMTC (massive Machine Type\\Communications)\end{tabular}} & \multicolumn{1}{l}{\begin{tabular}[c]{@{}l@{}}MBRLLC (mobile broadband\\reliable low latency communication), \\ mURLLC (massive URLLC), \\ HCS (human-centric services), \\ MPS (Multi-purpose 3CLS and\\energy services)\end{tabular}} \\ \bottomrule
\end{tabular}}
\end{table*}

Beyond the requirements of higher network performance, future 6G wireless networks are also anticipated to achieve ubiquitous intelligence, in which each network node can continuously learn from environment and intelligently adapt to network changes. Benefiting from the powerful capabilities of data analysis and modeling, Artificial intelligence (AI) technology is considered as an indispensable tool in future networks. Walid \cite{bib1} outlines the trends, applications and challenges for future 6G networks, and proposes that AI technology is critical to the paradigm shift of 6G from self-organizing networks (SON) to self-sustaining networks (SNN). AI-powered 6G SNNs can intelligently adapt its functions to different environments, thus providing a high quality of service under highly dynamic and complex environment. Additionally, network intelligence will be pushed at the network edge in future 6G networks, and it enables the wireless networks to run AI algorithms on edge devices to realize distributed autonomy. Yang \cite{bib2} proposed an AI-enabled intelligent architecture for 6G networks to realize data analysis, smart resource management and intelligent service provision. The author provides different AI-powered application examples under the proposed framework. For example, the author employs Principal Component Analysis (PCA) and Isometric Mapping (ISOMAP) to transform higher-dimensional data into lower-dimensional subspace to decrease the storage space, computing time and transmitting cost in dense networks.

However, the development of 6G technology will bring new challenges to the network operations and maintenance as well. First, intense connecting applications will increase the complexity of the network structure. It is predicted that by 2030, the number of users and devices in 6G networks will reach 17 billion with 60ZB data stored, far beyond the capacity of 5G. Second, in order to provide more reliable and lower-latency network services, 6G networks must meet more granular and stringent service-level agreements. In particular, operators need to quickly detect and diagnose abnormal patterns in the system. Data privacy and security, i.e., preventing disclosing sensitive information to unauthorized entities may increase the difficulty of applying AI algorithms to network management as well. For example, in terms of traditional machine learning methods, especially centralized machine learning (CML), raw data needs to be transmitted from local devices to centralized servers for processing and training, thus leading to compromised user privacy and security. Finally, more sensors and smart devices will be connected into future 6G networks, and the collected data can be heterogeneous with different formats, including images, text, time series and so on. According to whether the data is structured, the heterogeneous data can be divided into structured data, semi-structured data and unstructured data. In addition to this, the system usually collects KPI time series data, alarm event data, and log data to help the grasp of status in wireless networks. It may significantly help derive a more accurate analysis of faulty systems by utilizing the information of different types of data. This leads to the necessity of handling heterogeneous data in future networks. However, most existing RCA methods can only deal with homogeneous data, which means that the information in different data is not fully exploited. Therefore, how to effectively process and integrate heterogeneous data to achieve more efficient and accurate maintenance will be a challenge in 6G networks.

Today, mobile network operators already spend a large portion of their revenue on network maintenance and management, of which a significant portion is spent on fault diagnosis. Any system outage or downtime will potentially lead to a large revenue loss. It is foreseeable that the relevant investment will be even larger in 6G network maintenance. Therefore, effective anomaly detection (AD) \cite{activeIDS} and RCA are crucial for 6G wireless networks. 6G RCA differs from that of 5G in three aspects. First, 6G is expected to be more complex than 5G because of more connected smart devices, denser cell deployments, and increased diversities of various applications. The increased complexity and scale of 6G render the traditional RCA approach not directly applicable. Second, in 5G networks, RCA is typically carried out only after a fault or defect is detected, which is called reactive RCA. However, with the emergence of many time-sensitive applications and increased network complexity, \emph{real-time RCA} will become imperative for future 6G networks. This implies that the envisioned 6G networks require real-time proactive RCA to minimize the damage caused by system faults. Finally, as edge intelligence is envisioned to be a key building block of 6G wireless networks, for privacy protection, user data needs to be stored at the edge. Consequently, salable RCA method that can be implemented in a fully distributed manner is strongly desired. In summary, 6G RCA needs to deal with a wireless network with unprecedented \emph{scale and complexity} in \emph{real-time} and \emph{fully distributed} manner.   

The remainder of this paper is organized as follows. First, we devote two sections comprehensively introducing existing works on root cause analysis and causal discovery methods separately. Next, we discuss the existing causal discovery-based RCA methods and their limitations for application in future 6G. Then we elaborate on details of the proposed AI-empowered Net-RCA model. Finally, we conclude this paper.

\renewcommand\arraystretch{1.5}
\begin{table*}[!t]
\caption{Importance Levels of Key Network Requirement for 6G Use Cases (Taken from the ITU-T technical report) \label{tab:table2}}
\centering
\resizebox{\linewidth}{!}{
\begin{tabular}{c ccccc}
\bottomrule
\multirow{2}{*}{\textbf{Use Cases}}             & \multicolumn{5}{c}{\begin{tabular}[c]{@{}c@{}}\textbf{Importance Levels of Key Network Requirement}\\ (*1 to 3 are for relatively low; 4 to 6 are for medium; \\7 to 9 are for relatively high; and 10 means extremely demanding)\end{tabular}} \\ \cline{2-6} 
                                                & \multicolumn{1}{c}{\textbf{Bandwidth}}              & \multicolumn{1}{c}{\textbf{Latency}}              & \multicolumn{1}{c}{ \textbf{Artificial Intelligence} }              & \multicolumn{1}{c}{\textbf{Security}}             & \multicolumn{1}{c}{\textbf{Mobility}}             \\ \hline
Holographic Type Communication (HTC)            & \multicolumn{1}{c}{10}                              & \multicolumn{1}{c}{7}                             & \multicolumn{1}{c}{5}                        & \multicolumn{1}{c}{5}                             & 1                             \\ \cline{2-6}
Tactile Internet of Remote Operation (TIRO)     & \multicolumn{1}{c}{4}                               & \multicolumn{1}{c}{10}                            & \multicolumn{1}{c}{3}                        & \multicolumn{1}{c}{7}                             & 2                             \\ \cline{2-6}
Intelligent Operation Network (ION)             & \multicolumn{1}{c}{3}                               & \multicolumn{1}{c}{9}                             & \multicolumn{1}{c}{10}                       & \multicolumn{1}{c}{8}                           & 4                             \\ \cline{2-6}
Network and Computing Convergence (NCC)         & \multicolumn{1}{c}{5}                               & \multicolumn{1}{c}{8}                             & \multicolumn{1}{c}{5}                        & \multicolumn{1}{c}{5}                             & 3                             \\ \cline{2-6}
Digital Twin (DT)                               & \multicolumn{1}{c}{6}                               & \multicolumn{1}{c}{7}                             & \multicolumn{1}{c}{6}                        & \multicolumn{1}{c}{9}                             & 5                             \\ \cline{2-6}
Satellite-Terrestrial Integrated Network (STIN) & \multicolumn{1}{c}{5}                               & \multicolumn{1}{c}{6}                             & \multicolumn{1}{c}{2}                        & \multicolumn{1}{c}{7}                             & 10                            \\ \cline{2-6}
Industrial IoT (IIoT) with Cloudification       & \multicolumn{1}{c}{8}                               & \multicolumn{1}{c}{10}                            & \multicolumn{1}{c}{3}                        & \multicolumn{1}{c}{8}                             & 8                             \\ \bottomrule
\end{tabular}}
\end{table*}

\section{Root Cause Analysis Method}

\subsection{RCA}

In recent years, research activities have embraced AI technology advancements for network managements and operations, such as predictive maintenance (PdM). PdM is a condition-based maintenance technique, which is widely used and studied in various systems. Different from the traditional maintenance management, which mainly relies on the routine maintenance of system components and fast response to the unexpected faults, PdM is designed to perform the prediction and prevention of system failures based on the real-time monitoring of the system conditions. PdM can effectively reduce the number of unexpected failures, and improve the availability and reliability of systems. With the introduction of AI technologies, predictive maintenance is expected to play a greater role in the intelligent maintenance of future wireless networks.

\textbf{RCA} is an important research field that aims to maintain the reliability of a complex system. RCA can be loosely defined as the systematic process of discovering the root causes of abnormal system behaviors that may give rise to performance degradation of a complex system. RCA assumes that it is much effective to prevent and diagnose the underlying root causes of system failures than merely solve all problems that developed. By diagnosing the root causes of system anomalies, RCA can help engineers better understand system status and prevent system underperformance or costly system failures. There are three major types of RCA methods.

\subsection{Trace-based RCA}

Trace-based RCA algorithm aims to use the invocation information between nodes to infer the fault propagation of the system. This type of method is typically used in systems based on a service or microservice architecture, in which a request is achieved through the mutual calls of different services. Root cause anomalies may propagate through call paths, resulting in anomalies of multiple microservice nodes. TraceAnomaly \cite{bib4} is an RCA algorithm proposed for microservice systems, which comprehensively considers the response time and call path. Specifically, the algorithm first constructs multi-dimensional Service Trace Vectors (STVs) according to call paths under different response times, wherein each dimension of the vector represents a microservice on the call path. STVs are then input to a variational autoencoder (VAE) model, and the well-trained model is used to detect call-path anomalies. Finally, TraceAnomaly uses 3-sigma to perform anomaly detection on the response times of each dimension in anomalous call paths to identify the root cause. However, trace-based methods usually require call paths or topology information, which is difficult to obtain in many complex networks.

\subsection{Rule-based RCA}

Rule-based method analyzes root causes by mining association rules related to anomalous indicators in historical data, and this algorithm is widely used in different types of systems. In wireless network systems, the network performance is measured by many performance indicators, such as the downlink throughput, the delay and received signal strength indicator. The key quality indicators (KQIs) are used to characterize the quality of services (QoS) for user experience, and KPIs are used to monitor the coverage and traffic conditions of wireless networks. In order to ensure sound user experiences, network operators need to monitor the performance indicators and identify any abnormal behaviors of systems. In a wireless network, user experience may be influenced by the variations of network conditions, resulting in KQI anomalies. Yang \cite{bib5} proposed deep network analyzer (DNA), which is a big data analytics platform for RCA in mobile wireless networks and implemented on Apache Spark. By mining the association rules between anomalous KQIs and KPIs, DNA can identify the corresponding root causes. Specifically, DNA abstracts the RCA into two modules, which are the module of rule (fingerprint) learning and the module of anomaly detection and fingerprint matching. In the first module, the author employs a rare association rule mining (RARM) approach to find the association between KQIs and KPIs, and builds a fingerprint database by finding all rules from the historic data. When a KQI anomaly is detected, DNA compares the KPIs related to anomalous KQI against fingerprints in the database, and outputs a list of potential root causes. Wang \cite{bib6} proposed RCSF, a rule-based RCA method for microservice architecture systems. Different from TraceAnomaly, RCSF uses the frequent pattern mining (FPM) in call paths to mine associations between alarms and anomalous microservices. The author assumes that root cause microservices tend to appear more frequently than others because they may be invoked by every alarming business component. Therefore, RCSF applies FPM on all anomalous call paths related to anomaly alarms, and the most frequent subsequence is considered as the root cause. However, rule-based method requires a large amount of training data, and suffers from high computational complexity, which inevitably increase difficulties of their applications in large and complex networks.

\subsection{Graph-based RCA}

Graph-based RCA method aims to diagnose root causes by learning a dependency graph with multiple nodes. Jiang \cite{bib7} proposed the invariant graph (IG) model, which models the system by learning the invariant relationship of KPIs in normal time. Each edge in the graph represents an invariant relationship, which is called the invariant. The author uses the AutoRegressive eXogenous (ARX) model to establish invariants in the IG model. The ARX model is a regression model that uses the external variable as an additional input. It can be regarded as an invariant between the external variable and the endogenous variable. By learning all invariants, the system can be modeled as a complex IG model, and the edges in the graph are called invariant links. When a system gets anomaly, invariant links related to anomaly nodes will break, which is called broken links. Finally, the IG model outputs the top K most potential root causes by ranking the broken-link ratio of each node. Yong \cite{bib8} further developed an RCA algorithm for IG based on loopy belief propagation (LBP-IG). Different from the previous algorithm, this model builds a graph with both invariant and broken links to track the anomaly propagation of system. LBP-IG sets two initial features for each node, namely node abnormality degree and potential root cause degree, and assumes that anomalies between nodes will propagate with an initial probability. Under such an assumption, the author designs an LBP-based iterative algorithm. When the algorithm gets converged, LBP-IG outputs the most probable root cause based on the potential root cause degree. But the setting of initial parameters requires expert experience. And during the training of IG model, it needs to search all possible invariants, which takes high time consumption and limits its application in large-scale systems, such as 5/6G networks. MSCRED \cite{bib10} first obtains dependency graphs by calculating the correlation coefficients between a set of time series within different moving time windows. Then a CNN-based Autoencoder model is invoked to diagnose the root causes by calculating the error between the reconstructed output and the input.

In the communication networks, the systems typically collect a large amount of monitoring data through various sensors and devices, including KPI time series data, alarm event data and system log data. While the architecture of 6G may differ from its predecessors, analyzing such monitoring data to gain actionable insights plays an essential rule in RCA. Therefore, although 6G networks are expected to be deployed in 2030 and currently under development, the aforementioned RCA approaches, with proper modifications, may be applied to 6G as well. As an instance, RCSF \cite{bib6} extracts the anomalous periods in KPI time series as abnormal events, and employs frequent mining algorithms to analyze the associations between different types of alarms and KPI anomalies. Such an approach can potentially be employed to obtain the associations between alarm events and KPI time series anomalies in 6G. Therefore, with the introduction of more and more AI technologies, AI-based models will have greater application prospects in future 6G systems.

\begin{figure*}[!t]
\centering
\includegraphics[width=0.9\textwidth]{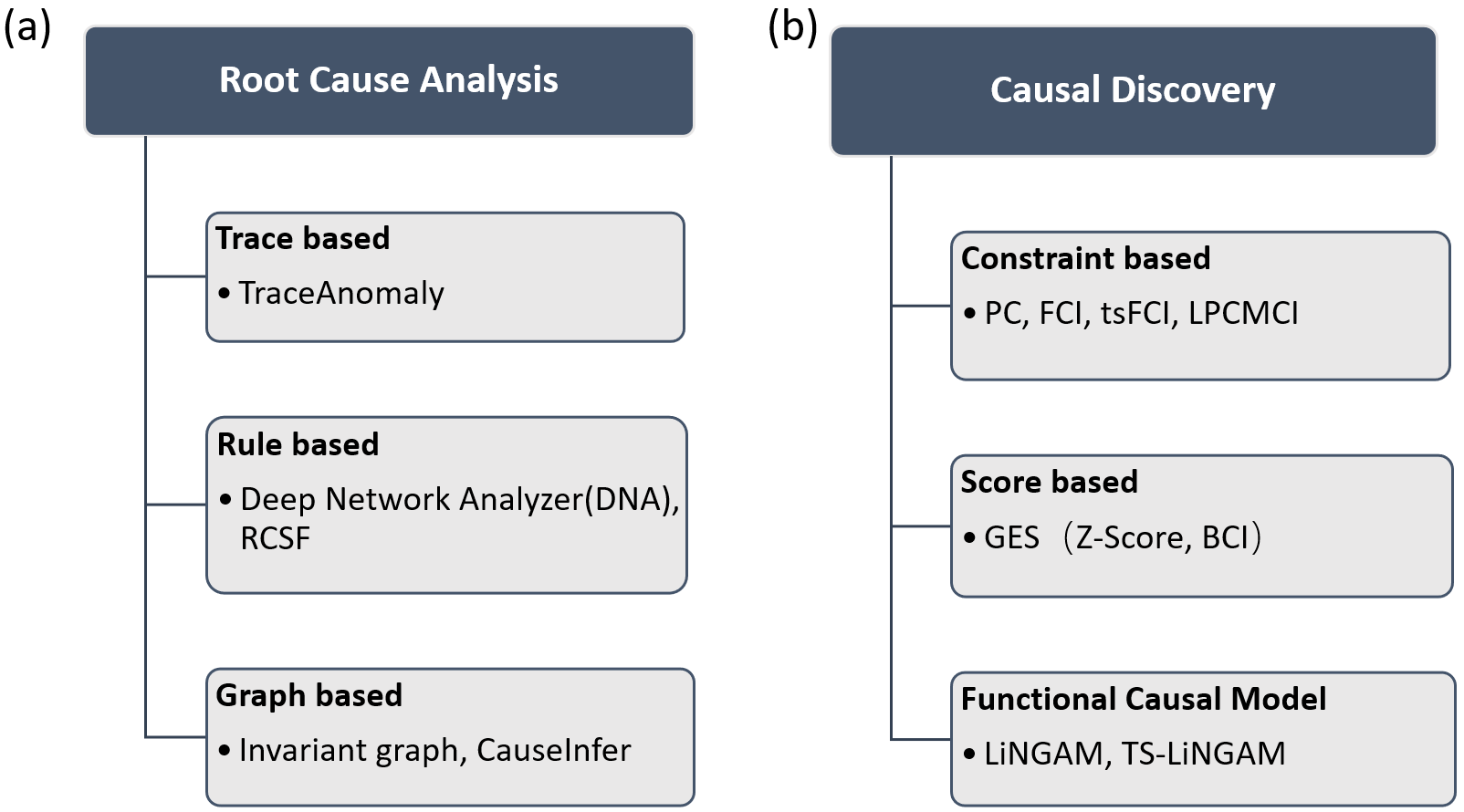}
\caption{Categorization of (a) Root cause analysis methods and (b) Causal discovery methods.}
\label{fig_2}
\end{figure*}

\section{Graph-based Causal Discovery Method}

\subsection{Causal Discovery}

Graph-based causal discovery aims to infer the causal structure by analyzing observational data, which is a fundamental study in various disciplines of science \cite{bib9}. In this section, we systematically investigate causal discovery methods. The traditional way to discover causality is to conduct a controlled trial. However, such experiments are often time-consuming or even impossible to carry out in reality. Therefore, researchers have proposed many methods to discover causal relationships from the historic data, and there are three major categories, as shown in Fig.2(b).

\subsection{Constraint-based Causal Discovery}

Constraint-based approaches aim to recover the underlying causal structure by exploiting the conditional independence (CI) relationships in the given data. These methods provide a search framework for causal discovery that can plug into different statistical models for testing CI between different data. However, due to the large combinatorial space to be searched, these methods have high time complexity. PC algorithm \cite{bib9} is a typical constraint-based method, which starts with a complete undirected graph and uses the significance level of the CI test as a threshold to determine whether the edges should be eliminated. Finally, the PC algorithm uses orientation rules to determine the causal direction. However, PC cannot discover causalities with time delay, thus limiting its applicability. Moreover, it assumes that there are no latent confounders, ie., unobserved common cause of the two measured variables in the observation data, which results in low accuracy. The FCI algorithm \cite{bib9} takes latent confounders into account during the CI test, and can discover asymptotically correct causal graphs. However, it may output Markov Equivalence Classes (MEC) graphs because of simple orientation rules. And it cannot be directly applied to time series data. Recently, J. Runge \cite{bib11} found that the low effect size of CI tests is the main reason for the low recall of these algorithms, and proposed LPCMCI for time series data, which addressed the above challenges and can discover causalities with time delay. Although LPCMCI has promising performance on synthetic data, high noise in real systems will greatly reduce the performance of the algorithm. And LPCMCI also requires a large number of CI tests, especially for large-scale systems, which incurs high computational cost.

\subsection{Score-based Causal Discovery}

Score-based approach provides a search framework for causal graphs, which starts with an empty graph, and iteratively adds possible directed edges or removes unnecessary links according to a defined score function, until the maximum score is obtained. During each step, the decision of whether to add or eliminate directed edges depends on the defined score function. Greedy equivalence search (GES) \cite{bib12} is a well-known score-based method that takes the quasi-Bayesian score, such as the Bayesian information criterion (BIC) or the Z-score of a statistical hypothesis test, as a score function. A general disadvantage of this kind approach is that it must traverse all potential edges, and thus it is of high computational complexity. Besides, the model performance largely depends on the selected score function.

\subsection{Functional Causal Models}

Functional causal models (FCMs) aim to discover causalities by fitting variables into a defined function. Variables that satisfy this function are considered to have causal relationships. TS-LiNGAM \cite{bib13} is a FCM for time series data, which assumes that an effect can be generated by a linear transformation on a cause plus additive non-Gaussian noise. By training on different pairs of variables, TS-LiNGAM can finally derive all causal relationships. However, these models rely on strong assumptions and are difficult to generalize.

\section{Causal Discovery For RCA}

Graph-based causal discovery method is capable of discovering underlying casual relationships. Besides, many causal discovery methods can provide interpretable results. Recently, many researchers apply causal discovery models in RCA to locate the root cause of abnormal patterns by constructing causal graphs.

G-RCA \cite{bib14} aims to introduce causality into the RCA algorithm, but it constructs causal graphs manually, and hence remains infeasible for large-scale dynamic systems. By applying causal discovery models, \cite{bib15} proposed CauseInfer, which is an RCA algorithm for microservice architecture systems. In this system, user instructions are implemented through different call paths of microservices, and each microservice is measured by a variety of indicators. Therefore, CauseInfer builds a two-layer hierarchical causal graph, namely, the service-based causal graph and the indicator-based causal graph. The former graph constructs a microservice-dependency graph by using the invocation relationships between microservices, aiming to locate the root cause at a service level. The latter graph is a causal graph of indicators in a microservice, which is constructed by conditional independence tests of PC algorithm, with the purpose of further diagnosing the root cause in an indicator level. CauseInfer constructed the indicator-based causal graph for each microservice with a different business function. By constructing such a two-layer causal graph, the algorithm can pinpoint the root cause by detecting anomaly of performance indicators in a faulty microservice along the causal paths. Specifically, once an anomaly is detected, the algorithm will traverse the causal graph by using the depth-first search algorithm. When CauseInfer finds an abnormal node without child abnormal nodes, the indicators corresponding to this node is regarded as a root cause. Through the ablation experiment, it is found that CauseInfer can pinpoint the root causes accurately with the employment of causal graphs. However, CauseInfer uses PC algorithm to learn causal graphs, which can be extremely unreliable in more complex networks. Besides, PC algorithm cannot discover causal relationships of time delay, the information of which will be useful to help engineers understand and model anomaly propagation paths. Furthermore, CauseInfer needs to construct causal graphs of indicators for all microservice nodes, and the computational complexity of methods will be unacceptable in large-scale networks.

Generally speaking, these models provide inspiration for how to combine the advantages of causal discovery and RCA, but there is a certain gap to directly apply them to future 6G networks. The intelligent operation and maintenance of future 6G networks require more accurate and lower-complexity algorithms, thus providing improved network service quality. Therefore, we propose an AI-empowered Net-RCA method in this paper.

\section{AI-empowered Net-RCA}
Traditional RCA methods typically determine the root causes of system failures based on a set of predefined rules provided by subject matter experts \cite{bib14}. However, such methods often show poor performance and require significant human efforts. Artificial intelligence technique can provide an effective means for decision making under uncertainty and thus offer a new approach for RCA in many scenarios. AI-based RCA methods can be broadly classified into two categories, including traditional machine learning (TML)-based RCA methods \cite{bib5,bib6,bib7,bib8} and neural network (NN)-based RCA methods. For example, DNA \cite{bib5} and RCSF \cite{bib6} employ FPM algorithms to obtain association rules with high confidence from the training data and subsequently apply the obtained rules to diagnose the root cause. \cite{bib7,bib8} use regression models to learn the correlation patterns between time series. Such correlation patterns are then fed into a machine learning model for root cause analysis. NN-based RCA algorithms have received widespread attention with the advance of deep learning techniques, including convolutional neural network (CNN), recurrent neural network (RNN), graph neural network (GNN). For example, TraceAnomaly \cite{bib4} and MSCRED \cite{bib9} employ NN models to learn correlation patterns among different time series. Such patterns can effectively assist root cause analysis in the next step.

Although many AI techniques have been used in RCA, most works have two common characteristics, which may limit their applications in 6G. On the one hand, the model building relies on expert domain knowledge or network topology. On the other hand, some methods have high time complexity and is not scalable, and thus is not applicable to large-scale systems. Additionally, to the best of our knowledge, employing AI-empowered RCA in wireless networks while considering causal relationships between anomalous indicators has not been systematically investigated before. Therefore, we mainly focus on how to perform accurate and efficient RCA in future 6G wireless networks by utilizing correlation analysis and causal discovery.

\subsection{Problem Definition}

The performance of a wireless network is measured via a collection of KPI time series. In this paper, we focus on root cause analysis of wireless network performance issues by analyzing these KPI time series. In particular, the RCA is defined as the process of identifying a subset of KPI time series that can be utilized to diagnose a root cause of the network performance degradation.

In future 6G networks, network performance is expected to be measured by more than thousands of performance indicators. The internal relationships between metrics are complex and difficult to be fully described by topological relationships or expert experience. Therefore, it is assumed that no or only limited prior knowledge is available in our problem to ensure that the proposed approach can be adapted to a wider range of scenarios. The problem of this paper is stated as follows. The indicators regularly collected from systems are multivariate time series, defined as $\mathbf{X}=\left\{\mathbf{x}_{1}, \mathbf{x}_{2}, \ldots, \mathbf{x}_{\mathrm{T}}\right\}$, where $\mathrm{T}$ is the length of $\mathbf{X}$. Each observation $\mathbf{x}_{t}$ is a D-dimensional vector at timestamp $t(t<T)$, denoted as $\mathbf{x}_{t}=\left[x_{t}^{1}, x_{t}^{2}, \ldots, x_{t}^{D}\right]$. Each dimension of the vector represents the value of different indicators. When system is anomalous at timestamp $t$, the behaviors of one or more indicators deviate from the normal patterns. Our proposed method will try to localize the top $\mathrm{N}$ indicators that are the most probable root causes of abnormal patterns.

\subsection{Proposed Method}

We present a Net-RCA method for future wireless networks without relying on any expert knowledge, named Temporally and Causally oriented RCA (TCoRCA). Our method can not only identify the system anomaly, but also diagnose the root cause efficiently and accurately. The workflow of TCoRCA is shown in Fig.3, which consists of two modules: a graphical network module and an RCA module.

\begin{figure}[!t]
\centering
\includegraphics[width=1.0\textwidth]{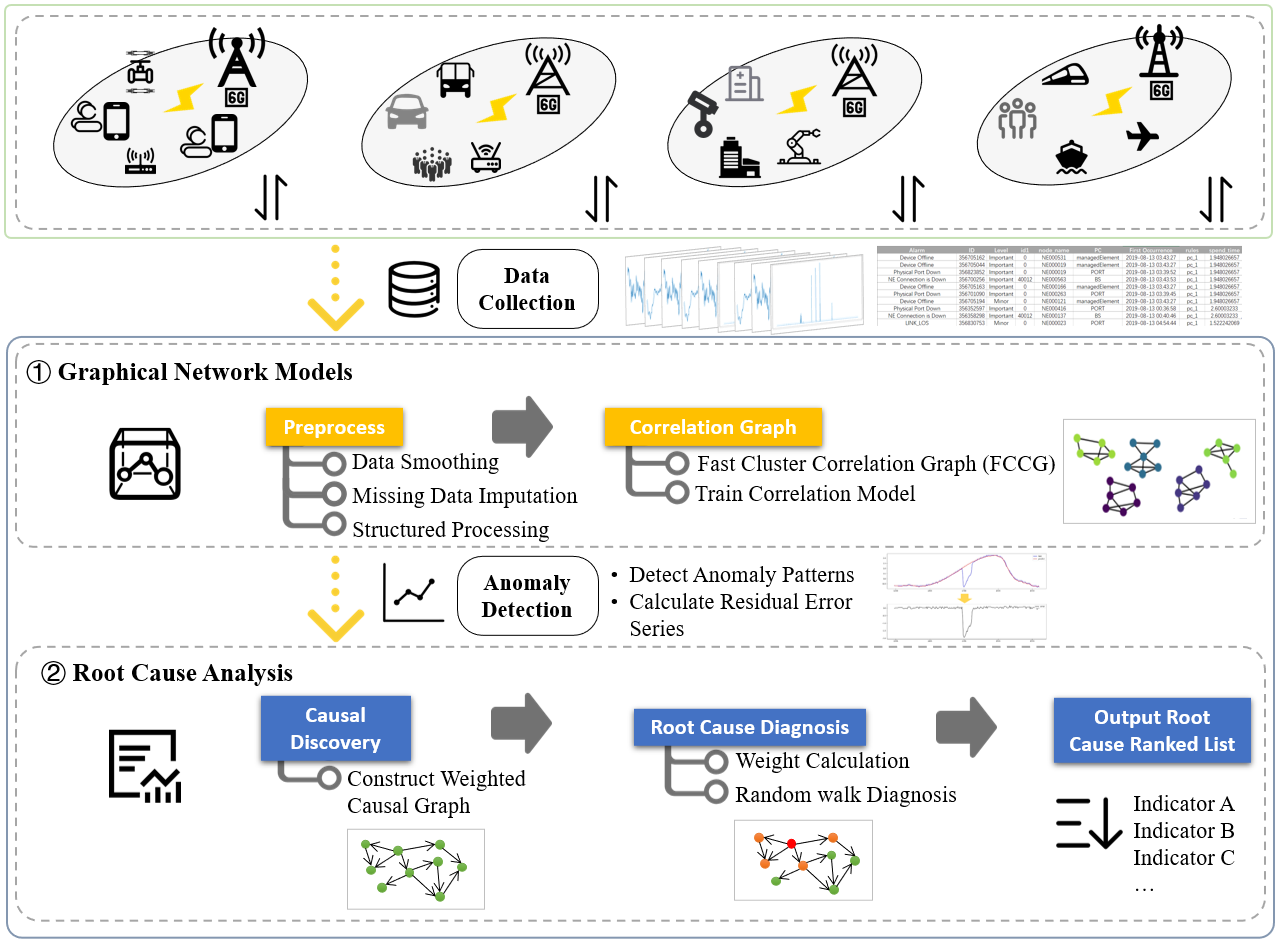}
\caption{TCoRCA Framework.}
\label{fig_3}
\end{figure}

In the module of graphical network, TCoRCA constructs a correlation graph consisted of multiple subgraphs to grasp the temporal dynamics of wireless networks, by analyzing the correlations between different indicators. First, TCoRCA preprocesses the input data, including data smoothing, missing data imputation and data structured processing. Then, we propose a fast cluster correlation graph (FCCG) method with lower time complexity $O(n \sqrt{n})$. FCCG employs invariants to establish the correlations between indicators. It iteratively runs two steps, including randomly selecting a target variable $x_i$ and clustering variables that shows strong invariant relationships with $x_i$, until all variables have been assigned to a cluster. The invariant relationships should hold at all times when the system is normal, and broken links can be used to detect anomalous patterns. When an anomaly is detected, the residual error series of anomalous indicators will be used as the input of RCA module. It is worth mentioning that both linear and nonlinear invariant relationships can be obtained via our algorithm.

In the proposed RCA module, we assume that causal relationships between various KPI time series play an important role in diagnosing root causes. In contrast to traditional methods based on correlation, causal discovery can capture more complex dependency patterns among more than two indicators, namely “conditional dependence” \cite{bib15}. This allows us to assess whether a change in one KPI will directly affect the distribution of another KPI. Such conditional dependence can effectively assist the root cause analysis.

Furthermore, the proposed Net-RCA method can be realized in a distributed manner based on the federated learning approach to better protect data privacy.

\subsection{Experiments}

We compare the performance of TCoRCA with threshold-based RCA, IG \cite{bib7} and LBP-IG \cite{bib8} models. The threshold-based RCA method diagnoses the root causes of system failures by comparing the measured KPI with a predefined threshold. To assess the performance of proposed RCA, we employ precision, recall and F1 score as the evaluation metrics.

We first present an example of proposed method on synthetic datasets, which are generated by trigonometric functions. By randomly selecting the frequency and time delay, we can obtain a set of time series. Then we randomly inject a pre-specified number of anomalies into these time series, which are considered to be the root causes. As shown in Fig. 4, the orange and green lines represent the performance of the two AI-based baseline methods, respectively. The yellow line corresponds to the performance of the threshold-based RCA method and the blue line shows the performance of the proposed algorithm. It is seen that TCoRCA achieves better performance than the baseline methods in all scenarios. Besides, the recall of our method remains relatively stable and is better, i.e., higher than 0.75. This is because TCoRCA diagnoses root causes by employing causal discovery method, which is more effective in discovering dependencies among KPI time series.

\begin{figure}[!t]
\centering
\includegraphics[width=3.5in]{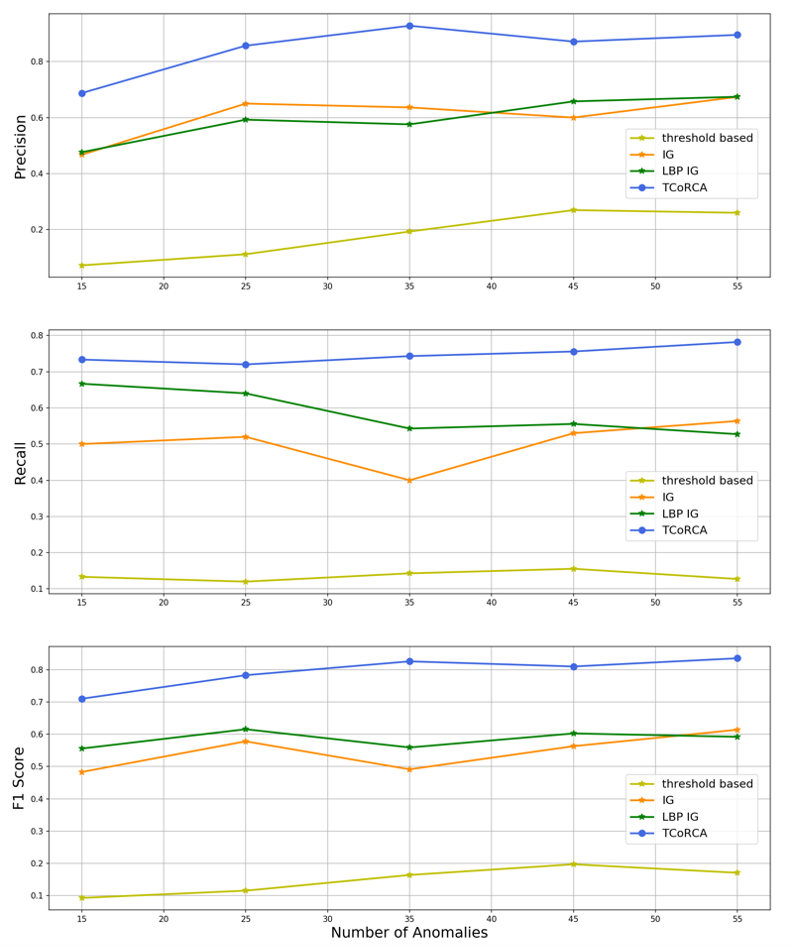}
\caption{Performance comparisons of different RCA methods on synthetic datasets.}
\label{fig_4}
\end{figure}

As another example, we investigate the performance of the proposed method on a real network traffic data. This dataset was collected from a communication network. Each cell is measured by a collection of performance indicators and the true root causes are provided by domain experts. The result shows that our algorithm outperforms the baseline method considerably. Compared with LBP-IG, the F1 score of our method improves from 0.5278 to 0.8117, and the recall increases by 66.48$\%$ i.e., from 0.5343 to 0.8895.

\section{Conclusion}

Since 6G is expected to be more reliable than 5G, preventive maintenance, in particular, proactive root cause analysis for network faults will become more important than ever. In this paper, we first survey three main approaches for root cause analysis in the literature. We then discuss a collection of causal discovery methods and how they can be applied to root cause analysis. We also present a novel AI-based RCA method. It is seen through experiments that the proposed approach achieves much better performance than the existing method. While 6G is still in its infancy, we expect Net-RCA will be a must for the 6G network operation and management, as it has done for legacy 4G and 5G systems. We hope this article will spur future research in this new and exciting area and pave the way for future 6G networks.

\bibliographystyle{abbrv}
\bibliography{main}

\raggedbottom
\pagebreak

\end{document}